\documentclass[letterpaper, 10 pt, journal, twoside]{IEEEtran}
\usepackage{graphicx}
\usepackage{paralist}
\usepackage[inkscapelatex=false]{svg}
\usepackage[shortlabels]{enumitem}
\usepackage{hyperref}
\usepackage{multirow}
\usepackage{colortbl}
\usepackage{tabulary}
\usepackage{etoolbox}
\usepackage{makecell}
\usepackage{tikz}
\usepackage{changepage}

\usepackage{amsmath}
\usepackage{mathptmx}

\usepackage{algorithm}
\usepackage[noend]{algpseudocode}
\usepackage{algcompatible}
\usepackage{lineno}
\usepackage{comment}
\usepackage{amssymb}
\usepackage{microtype}
\usepackage{bm}
\usepackage{colortbl}
\usepackage{cite}

\definecolor{control}{rgb}{0.1, 0.2, 0.5}
\definecolor{action}{rgb}{0.8, 0.1, 0.2}

\begin{document}

\title{Information-Theoretic Detection of Bimanual Interactions for Dual-Arm Robot Plan Generation}


\author{Elena Merlo, Marta Lagomarsino, and Arash Ajoudani
\thanks{Manuscript received: November 4, 2024; Revised: January 2, 2025; Accepted: March 5, 2025.
This paper was recommended for publication by Editor Júlia Borràs Sol upon evaluation of the Associate Editor and Reviewers’ comments.

This work was supported by the European Union Horizon Project TORNADO (GA 101189557). 
The authors are with Human-Robot Interfaces and Interaction (HRII) Laboratory, Istituto Italiano di Tecnologia, Genoa, Italy. Elena Merlo is also with the Dept. of Informatics, Bioengineering, Robotics, and Systems Engineering, University of Genoa, Genoa, Italy. Corresponding author's email: {\tt\footnotesize elena.merlo@iit.it}.

Digital Object Identifier (DOI): see top of this page.}
\vspace{-0.9cm}
}

\markboth{IEEE Robotics and Automation Letters. Preprint Version. Accepted March, 2025}%
{Merlo \MakeLowercase{\textit{et al.}}: Information-Theoretic Detection of Bimanual Interactions for Dual-Arm Robot Plan Generation}


\maketitle


\begin{abstract}
Programming by demonstration is a strategy to simplify the robot programming process for non-experts via human demonstrations. However, its adoption for bimanual tasks is an underexplored problem due to the complexity of hand coordination, which also hinders data recording. 
This paper presents a novel one-shot method for processing a single RGB video of a bimanual task demonstration to generate an execution plan for a dual-arm robotic system. To detect hand coordination policies, we apply Shannon’s information theory to analyze the information flow between scene elements and leverage scene graph properties. The generated plan is a modular behavior tree that assumes different structures based on the desired arms coordination. We validated the effectiveness of this framework through multiple subject video demonstrations, which we collected and made open-source, and exploiting data from an external, publicly available dataset. Comparisons with existing methods revealed significant improvements in generating a centralized execution plan for coordinating two-arm systems.
\end{abstract}

\vspace{-0.5cm}
\section{Introduction}
\IEEEPARstart{D}{espite} the growing evidence of the benefits robots bring to industrial, healthcare, and domestic settings, such as increased productivity and enhanced well-being \cite{lorenzini2023ergonomic}, robot programming remains the domain of expert programmers, hindering a wider adoption and use of such platforms. 

Programming by Demonstration (PbD) has shown to be effective in making robot programming more accessible and intuitive \cite{ravichandar2020recent}.
Through hands-on demonstrations, PbD leverages the natural social learning process of mimicking others \cite{calinon2007learning}.
Typically, when a person performs a manual task, both hands are engaged, either actively, such as when carrying a bulky object, or with one hand assisting the other in performing the primary work \cite{guiard1987asymmetric}. For example, when writing, one hand holds the paper steady while the other moves the pen. Similarly, when stirring a cup, one hand supports the cup while the other stirs.
In the literature, the problem of dual-arm manipulation programming is extensively treated \cite{smith2012dual}, however, the adoption of PbD methods for bimanual tasks remains underexplored. That is partly due to the complexity of capturing and analyzing coordination features between the hands and the challenges associated with recording data from demonstrations.

A widely used PbD strategy is Kinesthetic Teaching (KT), which allows a demonstrator to physically guide the manipulator within its workspace through haptic interaction \cite{calinon2007learning}. During this process, the robot's onboard sensors record its states, generating data that enables the replication of the demonstrated trajectories \cite{takano2016real}.
While KT simplifies the correspondence problem \cite{nehaniv2002correspondence}, the quality of the demonstration heavily depends on the smoothness and dexterity of the human user, which is particularly difficult to maintain during bimanual activities that require moving two arms simultaneously.
In \cite{reinhart2012representation, gribovskaya2008combining}, the authors managed to transfer bimanual skills, such as handling objects like tubes or large dice, by applying KT on both arms of small-sized humanoid robots, leveraging their compact form. 

To streamline the demonstration phase, robots can be instructed to replicate actions performed by an external person through a technique known as Observational Learning (OL) \cite{pauly2021seeing}. This method utilizes systems to capture human movements accurately. In \cite{liu2023birp}, human arms' movements and object displacements were recorded using Optitrack and analyzed offline to generate corresponding robot motions.
Sensor-equipped gloves have been additionally used in \cite{ureche2018constraints} to monitor even individual finger movements. 
Although practical markerless and RGB-camera-based motion capture systems have been explored \cite{fortini2023markerless}, the data they provide and process still lack the precision required for low-level PbD. 

\begin{figure*}[t!]
\centering
\includegraphics[width=1\linewidth]{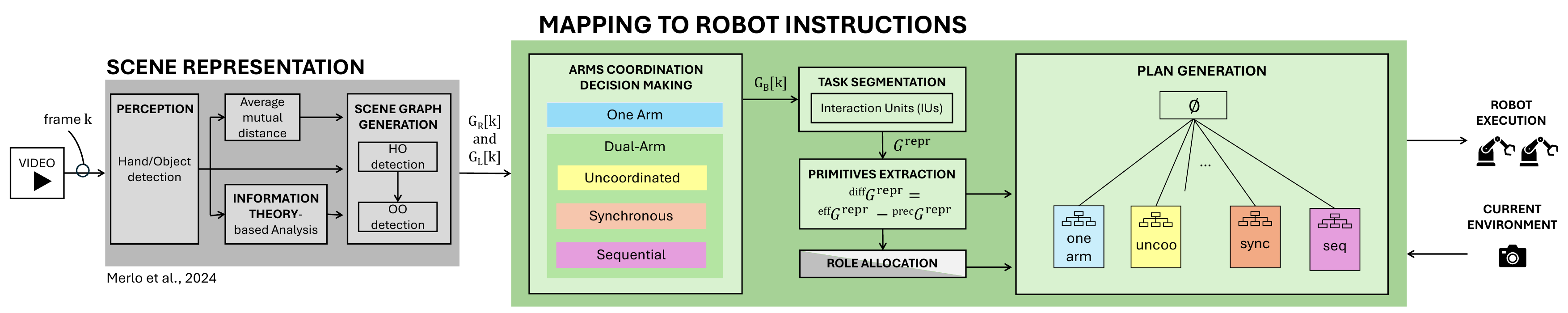}
\vspace{-0.7cm}
\caption{Framework overview. The first block maps each video frame $k$ into graphs $G_R[k], G_L[k]$ capturing task-relevant interactions for each hand, while the second block translates these into robot instructions, identifying arm coordination mode, extracting action sequences, and generating a dual-arm execution plan.}
\vspace{-0.5cm}
\label{fig:framework}
\end{figure*}

Recently, there has been growing interest in enabling robots to move beyond replicating demonstrated trajectories and gain \textit{higher-level} semantic understanding of human behavior \cite{ramirezamaro2017transferring}. This includes recognizing (i) the task goal, (ii) the sequence of required skills, and (iii) the environmental conditions necessary for successful execution \cite{zanchettin2023symbolic}. Such abstraction facilitates the generation of informed execution plans, improving robot decision-making \cite{diehl2021automated}. 
However, these advancements predominantly focus on unimanual actions. 
For instance, in \cite{dreher2020learning}, a graph-based structure models the hands-objects relations and trains a network to classify actions for each hand independently. While effective for unimanual tasks, this approach oversimplifies bimanual activities, as hand actions are often interconnected. 
Although many studies have explored hand coordination \cite{zollner2004programming, paulius2019manipulation, krebs2022bimanual}, less attention is paid toward methods for automatically identifying bimanual activities from human motion data to enhance dual-arm robotic manipulation.

This work presents a novel method to identify bimanual interactions and coordination from a single RGB video demonstration of a manipulation task and automatically generate a robot plan to map human actions into robot commands.
Compared to the state-of-the-art, our primary contribution lies in the application of Shannon’s Information Theory (IT) \cite{bossomaier2016introduction} to characterize bimanual activities. Our analysis of information flows enables us to recognize hand coordination policies, inspired by the taxonomy established in \cite{krebs2022bimanual}, and define the collaboration modalities for robotic arms.
Our second contribution consists of providing a direct mapping to a modular execution plan for a dual-arm system to enable it to replicate the demonstrated activity.
The proposed framework builds on our previous work \cite{merlo2024exploiting} to convert unimanual video demonstrations into time series of graphs that encode hand-object relations, including poses, distances, and shared information (see gray block in Fig. \ref{fig:framework}). 
The novelty compared to \cite{merlo2024exploiting} lies in extending this framework to accommodate bimanual activities, through the design of the green block in Fig. \ref{fig:framework}. 
Specifically, the first contribution involves analyzing the topology of the graphs encoding bimanual interactions and the informational content of their elements to determine the coordination mode of the two hands. Without this capability, the system would generate separate, uncoordinated plans for each arm, which are inadequate for tasks requiring close coordination and precise sequencing of the hands' action. 
The second contribution involves the automatic generation of a Behavior Tree (BT) \cite{colledanchise2018behavior} that guides the dual-arm system replica. Depending on arms' coordination, the BT adopts a diverse structure to manage the execution of coordinated movements and provide a solution for the role allocation problem, avoiding conflicting actions and ensuring logical execution.
The framework was tested on video demonstrations from two datasets and the retrieved task structure was compared with the one in \cite{dreher2020learning}.



\vspace{-0.2cm}
\section{Methods}
\vspace{-0.1cm}
Our algorithm consists of two main blocks: Scene Representation and Mapping to Robot Instructions, as illustrated in Fig. \ref{fig:framework}. The first block transforms each video frame into graphs capturing interactions between scene elements. We used IT-based measures to effectively extract the active part of the scene, i.e., the task-relevant elements, and create a compact and reliable scene representation. The second block translates these representations into an execution plan for the dual-arm system, identifying the arms' coordination mode and extracting the actions required. This pipeline enables intuitive robot programming by automatically generating structured instructions to replicate the observed demonstration.

\vspace{-0.3cm}
\subsection{Scene Graph Generation}
\vspace{-0.1cm}
\label{sec:graph_unimanual}

This section summarizes how hand activity in each frame $k$ is mapped into a scene graph $G[k]$ \cite{chang2023comprehensive}. In bimanual tasks, two independent graphs are generated per frame: $G_R[k]$ for the right hand and $G_L[k]$ for the left hand. A perception module captures the $6$D pose of each object $o \in O$ and of the hands $\{R, L\}$ in the scene, which constitute the nodes of the scene graphs. The positional data are then processed to evaluate the informational content of each signal and the amount of information shared between elements. 
Positional signals are also analyzed to obtain elements' relative distances. 
These outputs are used to connect nodes in $G_R[k]$ and $G_L[k]$. 
For more details on this process, refer to \cite{merlo2024exploiting}.

\subsubsection{Scene Graph Topology}
\begin{figure}
\centering
\includegraphics[width=0.75\linewidth]{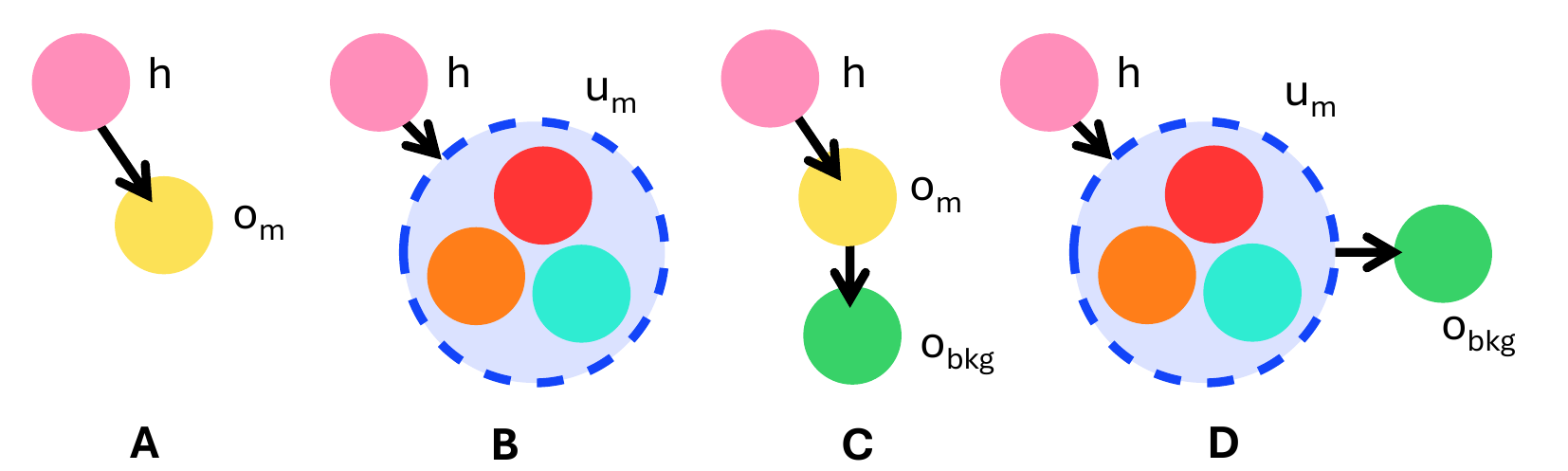}
\vspace{-0.2cm}
\caption{Possible topologies for $G_R$ and $G_L$ encoding single-hand interactions: (A) hand $h$ interacting with the manipulated object $o_m$; (B) hand manipulating a unity of three objects $u_m$; (C) interaction between $o_m$ and a static background object $o_{bkg}$; (D) $u_m$ interacting with $o_{bkg}$.}
\vspace{-0.6cm}
\label{fig:topologies}
\end{figure}

Since graphs encode only the active part of the scene, $G_R[k]$ and $G_L[k]$ can assume one topology among those shown in Fig. \ref{fig:topologies}.
In the graph, there is always a hand node ($h$), representing the right hand in $G_R$ and the left hand in $G_L$. 
The hand node can be connected to an object node $o_m$ (as seen in topologies A and C). This signifies a hand-object interaction ($HO$), indicating they share information during manipulation. Alternatively, the hand might interact with a unity of objects $u_m$ (as shown in topologies B and D). In these cases, the hand shares information with multiple objects simultaneously, such as when manipulating assembled objects or a box containing several items.
The object or unity being manipulated can also interact with another stationary object in the background $o_{bkg}$, as shown in topologies C and D. This static object-object interaction ($OO$) occurs when the information flow between the two elements reaches a state of equilibrium. An example is placing a pen inside a case, leading to the $OO$ between the pen and the case.
The oriented edges show that the graph generation process starts from the hand, which is the active element causing changes in the environment, and each edge contains the relative position between the element at the tail w.r.t. the element at the tip.

\subsubsection{Information Theory Metrics in Manipulation Task}
To determine if two elements in the scene captured by frame $k$ are sharing information (e.g., a hand and an object to establish a $HO$), we use \textit{Mutual Information} (MI). This measure, derived from Shannon's IT, is applied to positional signals $X$ and $Y$ of a pair of scene elements, considering a time interval $w$ centered in $t^*$, instant when frame $k$ was taken. $MI(X(t^*):Y(t^*))$ is defined as:
\vspace{-0.2cm}
\begin{equation}
\label{eq:mi}
MI(X(t^*):Y(t^*)) = \sum_{x \in \Omega_x} \sum_{y \in \Omega_y} p_{xy}(x,y) \cdot \log_{2}{\frac{p_{xy}(x, y)}{p_x(x)p_y(y)}},
\end{equation}
where $x$ and $y$ are discrete measurements of $X$ and $Y$ from $\Omega_x$ and $\Omega_y$, respectively, occurred within $w$; $p_x(x)$ represents the probability of $X$ taking value $x$, $p_y(y)$ of $Y$ taking value $y$, and $p_{xy}(x,y)$ is the joint probability of $X$ and $Y$ taking values $x$ and $y$ simultaneously.
If $MI(X(t^*):Y(t^*)) = 0$, it indicates that in $k$ the two elements are moving independently; otherwise, they are jointly moving. To recognize a manipulated unity $u_m$, we measure the shared information among the objects involved using \textit{Co-Information}, which extends MI to multiple variables \cite{bell2003co}.
This approach provides a more reliable alternative to contact estimation or distance measurements, preventing interaction misidentification based only on proximity. Moreover, it captures motion dependencies more effectively than velocity analysis due to a quantitative assessment of information flow, even with less accurate or noisy perception data \cite{merlo2024exploiting}.
Note that the value of MI between a hand and the manipulated object during a $HO$ is stored in the graph as an attribute of the corresponding edge.
Finally, to identify a static $OO$, we monitor the informational content of the distance between $o_m$ and $o_{bkg}$ by computing its entropy over time $H(\overline{d}_{o_m,o_{bkg}}(t))$. A drop suggests the objects' relative position is stabilizing, thus their interaction is relevant.

\vspace{-0.3cm}
\subsection{Arms Coordination Decision Making}
\vspace{-0.1cm}
\label{sec:arms_coordination}
The first step towards generating the execution plan is recognizing the roles the two hands played in the human demonstration by analyzing the relations between the information flows originating from the two hands during bimanual tasks. This allows us to design the plan to reproduce the same coordination modality between the two robotic arms.
The defined coordination modalities cover the majority of bimanual tasks, as supported by the established taxonomy in \cite{krebs2022bimanual}.

The coordination mode is automatically detected by inspecting the topology of $G_B$, the bimanual graph produced by the combination of $G_R$ and $G_L$. 
Each scene graph $G$ is defined as a 3-tuple $G = (V, R, E)$, where $V$ is the set of nodes, $E$ is the set of edges, and $R$ is the set of relations associated with each edge. Thus, $G_B$ can be expressed as the union of $G_R$ and $G_L$, i.e., $G_B = G_R \cup G_L$, where $V_B = V_R \cup V_L$, $E_B = E_R \cup E_L$, and $R_B = R_R \cup R_L$. Since $G_B$ is associated with a specific coordination mode, denoted as $c$, we can express $G_B = \{V_B, R_B, E_B, c\}$.
In Table \ref{tab:bim_top}, we displayed all the possible topologies for $G_B$, and we detailed their features and meaning. 
The node connected to the hand can represent a single object $o_m$ or a manipulated unity $u_m$. Dashed lines indicate optional edges.
The last column reports the coordination modality $c$ relative to each case.
%
%
%
%
\begin{table}[t]
\caption{$G_B$ features and relative coordination mode}
\label{tab:bim_top}
\centering
\begin{tabular}{|c||c|c|c|}
\hline
\textbf{Topology} & \textbf{Features} & \textbf{Description} & \textbf{$c$}\\
\hline
\makecell{\textbf{$\alpha$ / $\beta$} \\
    \centering \includegraphics[width=0.23\linewidth]{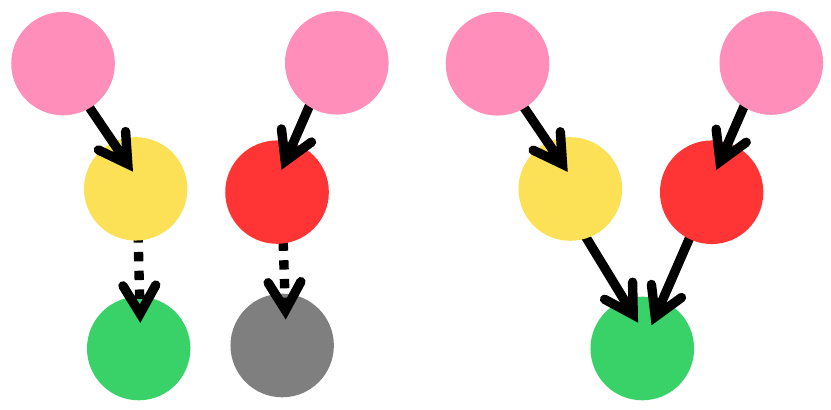}}
& \makecell{no nodes \\ in common / \\ same $o_{bkg}$} & \makecell{\scriptsize hands are independent \\ \scriptsize and do not influence each other \\ \scriptsize (e.g., transporting two glasses, \\ \scriptsize one with each hand / relocating \\ \scriptsize the two glasses on a shelf)} & \cellcolor{yellow!50}\makecell{\rotatebox{90}{\textbf{Uncoordinated }}} \\
\hline
\makecell{\textbf{$\gamma$} \\
\centering \includegraphics[width=0.09\linewidth]{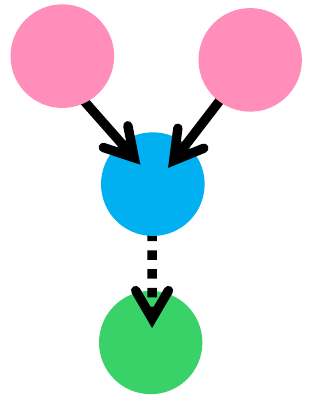}} & \makecell{same $o_{m}$ \\ (and \\ same $o_{bkg}$) } & \makecell{\scriptsize hands work together \\ \scriptsize  to manipulate the same object \\ \scriptsize 
 (e.g., moving a box)} & \cellcolor{orange!50}\makecell{\rotatebox{90}{\textbf{Synchronous }}} \\
\hline
\makecell{\textbf{$\delta$ / $\eta$} \\
\centering \includegraphics[width=0.23\linewidth]{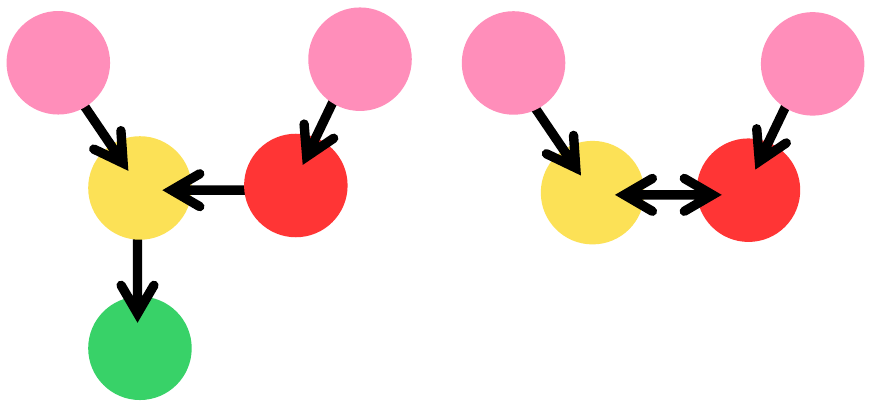}}
& \makecell{$o_{mL} = o_{bkgR}$ \\ or / and \\vice versa} & \makecell{\scriptsize the motion of one hand follows \\ \scriptsize (as in a sequence) or depends \\ \scriptsize  on the motion of the other \\ \scriptsize (e.g., holding a cup and \\ \scriptsize stirring with a spoon)} & \cellcolor{violet!50}\makecell{\rotatebox{90}{\textbf{Sequential }}} \\
\hline
\end{tabular}
\vspace{-0.4cm}
\end{table}
Regarding topology $\delta$ / $\eta$, we add that the two manipulated objects serve different roles: the reference object ($o_{ref}$) provides a reference for the dominant object ($o_{dom}$). For instance, when holding a cup with one hand and stirring it with a spoon using the other, the cup is $o_{ref}$, while the spoon is $o_{dom}$: the spoon's stirring action is indeed constrained by the cup's position.
To determine $o_{dom}$, we compare right hand-object ($RO$) and left hand-object ($LO$) $MI$. The higher one indicates which hand manipulates $o_{dom}$. 

If one hand is inactive, its corresponding graph will be empty. In this case, $G_B$ will represent the activity of only the active hand, and $c$ reports a \textit{one arm} activity, i.e., just one arm will be involved in the robot execution.



\begin{figure} [b]
\centering
\vspace{-0.5cm}
\includegraphics[width=0.85\linewidth]{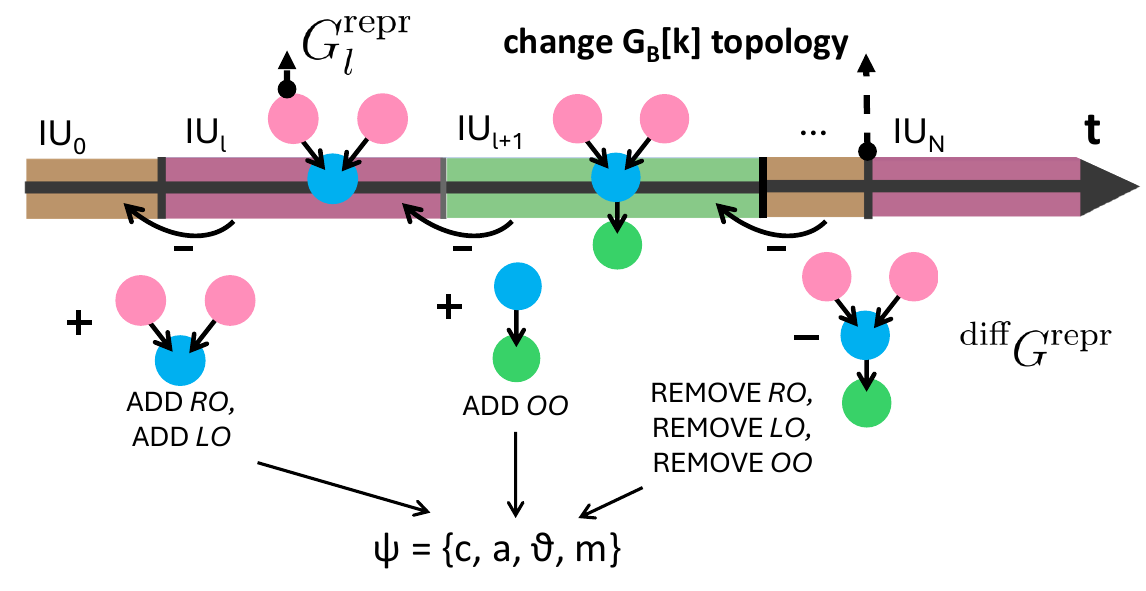}
\vspace{-0.3cm}
\caption{Segmentation of a \textit{synchronous} bimanual activity. Each retrieved IU is represented by its $G^\text{repr}$. Eq. \ref{eq:gdiff} is applied to adjacent $G^\text{repr}$ obtaining $^\text{diff}G^\text{repr}$, which includes sub-graphs encoding new (positive sign) and ended (negative sign) interactions. Each sub-graph is mapped into high-level primitives $\psi$.}
\label{fig:segm_bi}
\end{figure}

\alglanguage{pseudocode}
\begin{algorithm} [!b]
\small
\caption{\small High-Level Primitive $\psi$ Extraction} \label{algo:primitives_extr}
\begin{algorithmic}[1]
\State P $\gets \{\}$ {\scriptsize \Comment{\emph{list of primitives}}}
\For{$l \in$ [$1$, $\text{tot}_{IU}$]}
    \State $^\text{diff}G^\text{repr} \gets$ Eq. \eqref{eq:gdiff}   
    \If{$^\text{diff}G^\text{repr}$ reports a new RO / LO interaction}
        \State $a_{\psi_1} \gets \texttt{move}$
        \State $\theta_{\psi_1} \gets T^{o_m}_{R/L}$ {\small \Comment{\scriptsize \emph{hand-manipulated object relative pose}}}
        \State $a_{\psi_2} \gets \texttt{grasp}$
    \ElsIf{$^\text{diff}G^\text{repr}$ reports an ended RO / LO  interaction}
        \State $a_{\psi} \gets \texttt{release}$ 
    \ElsIf{$^\text{diff}G^\text{repr}$ reports a new OO interaction}
        \State $a_{\psi} \gets \texttt{move}$ 
        \State $\theta_{\psi} \gets T^{o_{bkg}}_{o_m}$ {\small \Comment{\scriptsize \emph{manipulated-background object relative pose}}}
    \EndIf
\EndFor
\State Add all $\psi$ to P
\end{algorithmic} 
\end{algorithm}

\vspace{-0.4cm}
\subsection{Task Decomposition}
\vspace{-0.1cm}
\label{sec:segm_alloc}
A second step in the plan generation process involves exploiting the temporal variations of $G_B$ to trace the sequence of actions performed. Specifically, we first segment the demonstrated task into Interaction Units (IUs), which are temporal blocks where the interactions between scene elements remain consistent. Then, we analyze the changes between couples of adjacent IUs to retrieve which action causes the transition.

To operate, we extract a representative graph $G^\text{repr}_l$ for the $l$-th IU, which encodes the topology of all graphs within that IU. 
This graph is selected based on $c$ and $MI$ values to capture the relative poses of the scene elements at key moments, like grasp initiation or final object-object configuration. 
To compare adjacent IUs, we perform a difference operation on $G^\text{repr}_l$:
\begin{equation}
\label{eq:gdiff}
    ^\text{diff}G^\text{repr} = \, 
    ^\text{eff}G^\text{repr} - \, 
    ^\text{prec}G^\text{repr} = \,
    G_{l+1}^\text{repr} - G_{l}^\text{repr},
\end{equation}
with $G_{l+1}^\text{repr}$ representing the effects of hands' action and $G_l^\text{repr}$ representing the preconditions. 
$^\text{diff}G^\text{repr}$ contains the sub-graphs corresponding to the new $RO$, $LO$, $OO$ interactions (positive sign) and/or those ended (negative sign).
Each sub-graph is translated into one or a combination of high-level primitives, which represent robot actions required to replicate the corresponding IU variation caused by the hand movements. Each action is expressed as a 4-tuple $\psi = \{c, a, \theta, m\}$. Here, $c$ is the coordination mode, $a$ is the action type for the robotic arm (e.g., \texttt{move}, \texttt{grasp}, or \texttt{release} \cite{guha2013minimalist}), $\theta$ includes action parameters, and $m$ identifies which hand executed the action for role allocation.

Fig. \ref{fig:segm_bi} graphically depicts how the segmentation process works for a bimanual \textit{synchronous} activity. The first difference between the retrieved $G^{\text{repr}}$ marks the beginning of new $RO$ and $LO$. Each $HO$ translates into a pair of primitives, $\psi_1$ and $\psi_2$. As shown in lines 4-7 of Algorithm \ref{algo:primitives_extr}, $a_{\psi_1}=$ \texttt{move} and $a_{\psi_2} =$ \texttt{grasp} reflect the approach and the grasp of $o_m$, respectively. One of the parameters in $\theta_{\psi_1}$ is the observed relative pose between the hand and $o_m$, represented as a transformation matrix. The second variation between IUs is due to a new $OO$ mapped into a primitive $\psi$ with $a_{\psi} =$ \texttt{move}, and in $\theta_{\psi}$, $T^{o_{bkg}}_{o_m}$ encoding the observed relative pose between $o_m$ and $o_{bkg}$ is included (see lines 10-12). Finally, the last difference indicates the end of $RO$ and $LO$. Each ending $HO$ translates into a primitive for the release of $o_m$ (lines 8-9). The end of $OO$ is not treated. All the generated primitives are stored in a list $P$. 

\vspace{-0.3cm}
\subsection{Plan Generation}
\label{sec:plan}
Once we have the sequence of actions to perform and the roles the two arms should take for proper coordination, we finally generate the execution plan in the shape of a BT.

A BT is a directed, rooted tree structure used for managing execution flows in robotics. It comprises internal nodes for control logic and leaf nodes for actions or conditions. The root node periodically sends a \textit{tick} signal down to its children, which execute and return a status (\textit{SUCCESS}, \textit{FAILURE}, or \textit{RUNNING}) based on their outcomes. 
BTs include four standard types of control nodes (sequence, fallback, parallel, decorator), which handle the call to their children according to diverse policies as detailed in Table \ref{tab:ctrl_nodes}, and two types of execution nodes (condition, action), whose symbols and return status are listed in Table \ref{tab:exe_nodes}. 
\begin{table}[b!]
    \centering
    \vspace{-0.3cm}
    \caption{BT Control Nodes and relative Use Cases}
    \begin{tabular}{|c|c|c|c|}
        \hline
        \textbf{Type} & \textbf{Symbol} & \textbf{Children Call} & \textbf{Use Case} \\ \hline
        Sequence   & $\rightarrow$   & \makecell{One after another \\ (sequential)}   & \makecell{ Ordered actions (e.g., grasp \\ \textit{if success then} move)} \\ \hline
        Fallback    & ?  & \makecell{One after another \\ (sequential)}   & \makecell{Selector behavior (e.g., \\ try primary instruction \\ \textit{if fail then} try alternative)}  \\ \hline
        Parallel    & $\begin{array}{c} \rightarrow \\ [-0.5em] \rightarrow \end{array}$  & \makecell{All at once \\ (concurrent)} & \makecell{Simultaneous actions \\ (e.g., move \textit{both} the arms \\ to different targets)} \\ \hline
        Decorator   & $\Diamond$   & Custom   & Custom  \\ \hline
    \end{tabular}
    \label{tab:ctrl_nodes}
\end{table}
\begin{table}[h]
    \centering
    \caption{BT Execution Nodes and Return Status}
    \begin{tabular}{|c|c|c|c|c|}
        \hline
        \textbf{Type} & \textbf{Symbol} & \textbf{Success} & \textbf{Failure} & \textbf{Running} \\ \hline
        Action      & $\square$   & \makecell{Upon \\ completion}   & \makecell{Impossible to \\ complete}   & \makecell{During \\ execution} \\ \hline
        Condition   & $\bigcirc$   & \makecell{True}  & \makecell{False}  & \makecell{Never} \\ \hline
    \end{tabular}
    \label{tab:exe_nodes}
    \vspace{-0.2cm}
\end{table}
\\
To automatically create the BT for replicating human activities, the ordered primitives in $P$ are analyzed one by one.
The first step involves checking the value of $c_{\psi}$ to identify the coordination mode. 
Whenever $c_{\psi}$ changes, a new control node is added to the root node, which starts a new subtree (referred to as $sBT$) containing the instructions required for that specific coordination mode. These subtrees are named according to $c_{\psi}$ ($sBT_{one-arm}$, $sBT_{uncoo}$, $sBT_{sync}$, and $sBT_{seq}$) and assume different structure.
Then, the value of $a_{\psi}$ is considered.
In case $a_{\psi} = \texttt{grasp}$ or $a_{\psi} = \texttt{release}$, $\psi$ is converted into an action node which handles commands for closing or opening the robot gripper, respectively, and is added as a child node to the initiated $sBT$.
When $a_{\psi} = \texttt{move}$, a further $sBT$ (namely, a sub-subtree) is generated and attached to the initiated structure, including the action node handling the trajectory computation and execution. 
The value of $m_{\psi}$ determines to which of the two arms (named $arm_x$ and $arm_y$) the action node is allocated. 
In the graphical BT representations, we indicate action nodes assigned to $arm_x$ with dark gray blocks and those to $arm_y$ with light gray blocks. 
In the following, we describe the $sBT$ corresponding to each coordination mode.

\subsubsection{One arm Subtree}
\begin{figure}
\centering
\vspace{-0.1cm}
\includegraphics[trim={0.5cm 0 0.5cm 0}, width=0.9\linewidth]{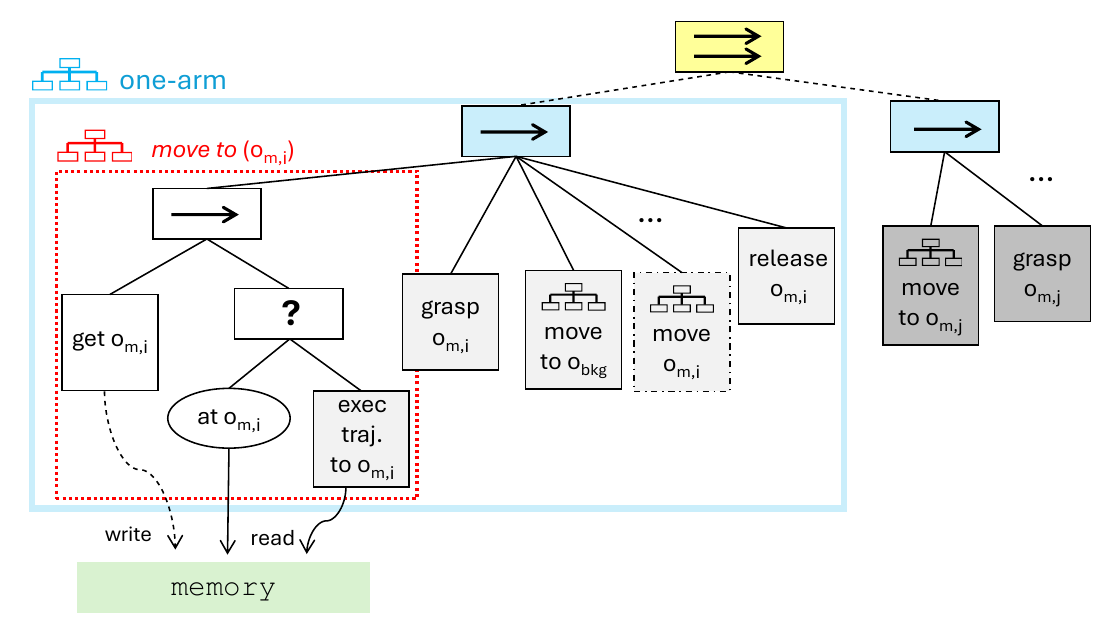}
\vspace{-0.4cm}
\caption{Subtree for \textit{uncoordinated} dual-arm activities, detailing the structure of \textit{one arm} subtree in the blue box. The subtree handling movements towards a target object is detailed as well, in the red box.}
\vspace{-0.5cm}
\label{fig:one_arm}
\end{figure}

The $sBT_{one-arm}$ has as root a sequence node.
The typical structure for this subtree is illustrated within the blue box of Fig. \ref{fig:one_arm}. 
One arm is commanded to reach and grasp $o_{m,i}$, move towards $o_{bkg}$, interact with $o_{bkg}$ following a specific motion pattern, and release $o_{m,i}$. Let us consider a wiping task where the arm is instructed to use a sponge to clean a surface: the robot first grasps the sponge, moves it close to the surface, and then executes the wiping motion based on the specific demonstrated pattern. After completing the task, the sponge is returned to its original position.
As shown, $sBT_{\textit{move to $o_{m,i}$}}$, framed in the red box, acquires and stores the pose of $o_{m,i}$ (the sponge in our example). The \textit{at $o_{m,i}$} condition node uses this data to verify that the end-effector is not already at the target before initiating the trajectory. The \textit{execute trajectory to} $o_{m,i}$ action node uses the stored pose, along with $T^{o_{m,i}}_{h} \in \theta_{\psi}$, to generate and follow the required trajectory, as detailed in \cite{merlo2024exploiting}. 
A similar subtree $sBT_{\textit{move to $o_{bkg}$}}$ is created to handle $o_{m,i}$ relocation movement (in our case, moving the sponge close to the surface to be cleaned), while $sBT_{\textit{move $o_{m,i}$}}$ handles the wiping motion. The dashed border around this node indicates that our pipeline currently recognizes complex motions like wiping, and the execution plan structure is set up to include them. However, these movements must be learned separately and then integrated in the BT.


\subsubsection{Uncoordinated Subtree}
In the case of \textit{uncoordinated} activities, where the two manipulators should perform different and unrelated tasks concurrently, the $sBT_{uncoo}$ consists of a father parallel node with two $sBT_{one-arm}$ as children: one containing instructions for $arm_x$ and the other for $arm_y$ (see blocks colored with the two gray shades in Fig. \ref{fig:one_arm}). In this case, $arm_x$ manipulates object $o_{m,i}$, while $arm_y$ handles $o_{m,j}$, without interference. The parallel node allows concurrent execution by two arms but does not ensure perfect synchronization, as this is unnecessary for the bimanual activity under consideration.
In the specific scenario where one or both hands hold an object without moving it, the corresponding $sBT_{one-arm}$ will feature a \textit{keep grasp} action node, directing the arm to maintain its grip.

\subsubsection{Synchronous Subtree}
\begin{figure}
\centering
\includegraphics[trim={0.5cm 0 0.5cm 0}, width=0.65\linewidth]{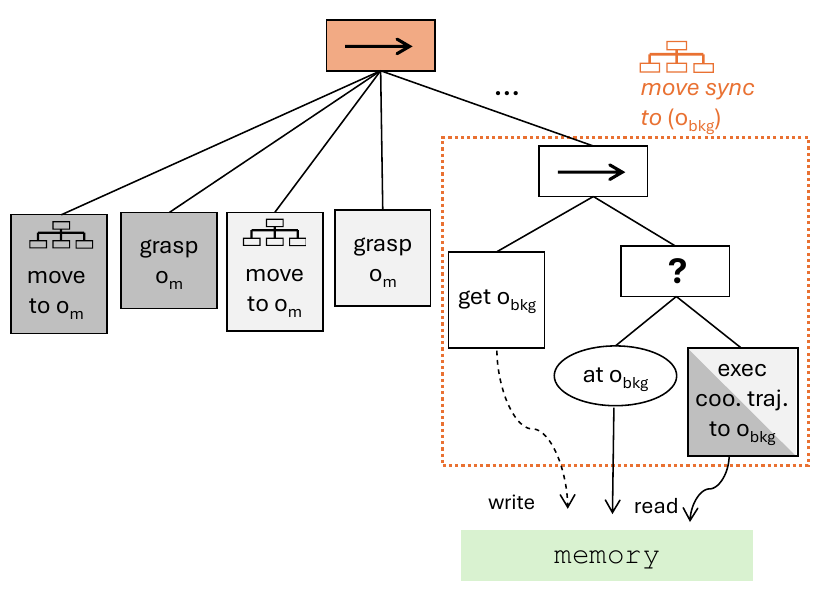}
\vspace{-0.4cm}
\caption{Subtree for a \textit{synchronous} dual-arm activity.}
\vspace{-0.6cm}
\label{fig:sync}
\end{figure}

For a dual-arm \textit{synchronous} activity, like relocating a tray with both hands, the $sBT_{sync}$ is structured as illustrated in Fig. \ref{fig:sync}. A single sequence node manages the commands for both arms to reach and grasp $o_m$ at distinct grasping points, replicating the human demonstration. The color-coded blocks, in two shades of gray, always indicate the allocation. However, the peculiarity of $sBT_{sync}$ consists in the design of the action node \textit{exec coordinated trajectory to $o_{bkg}$} within the $sBT_{\textit{move sync to $o_{bkg}$}}$, which commands both the arms to move coordinately.
This action node receives the current pose of $o_{bkg}$ relative to the camera frame ($T^{cam}_{o_{bkg}}$) and the desired final pose of $o_m$ in $\theta_{\psi}$ as $T^{o_{bkg}}_{o_m}$. The target pose for $o_m$ is computed by multiplying these matrices, and a trajectory is generated for $o_m$ to reach this target pose. Each trajectory point is mapped to each arm reference frame, creating two rigidly constrained trajectories that maintain a fixed distance between the end-effectors during manipulation. These trajectories are sent to the respective controllers in a synchronized, point-by-point manner.
Unlike the typical \textit{exec trajectory to $o_{bkg}$} action node, where the trajectory for $o_m$ is planned directly in the elected robot's base frame, here, it is initially planned in the camera frame and then projected onto each arm base.

\subsubsection{Sequential Subtree}
\begin{figure} [t]
\centering
\begin{adjustwidth}{-0.5cm}{-0.1cm}
\includegraphics[width=\linewidth]{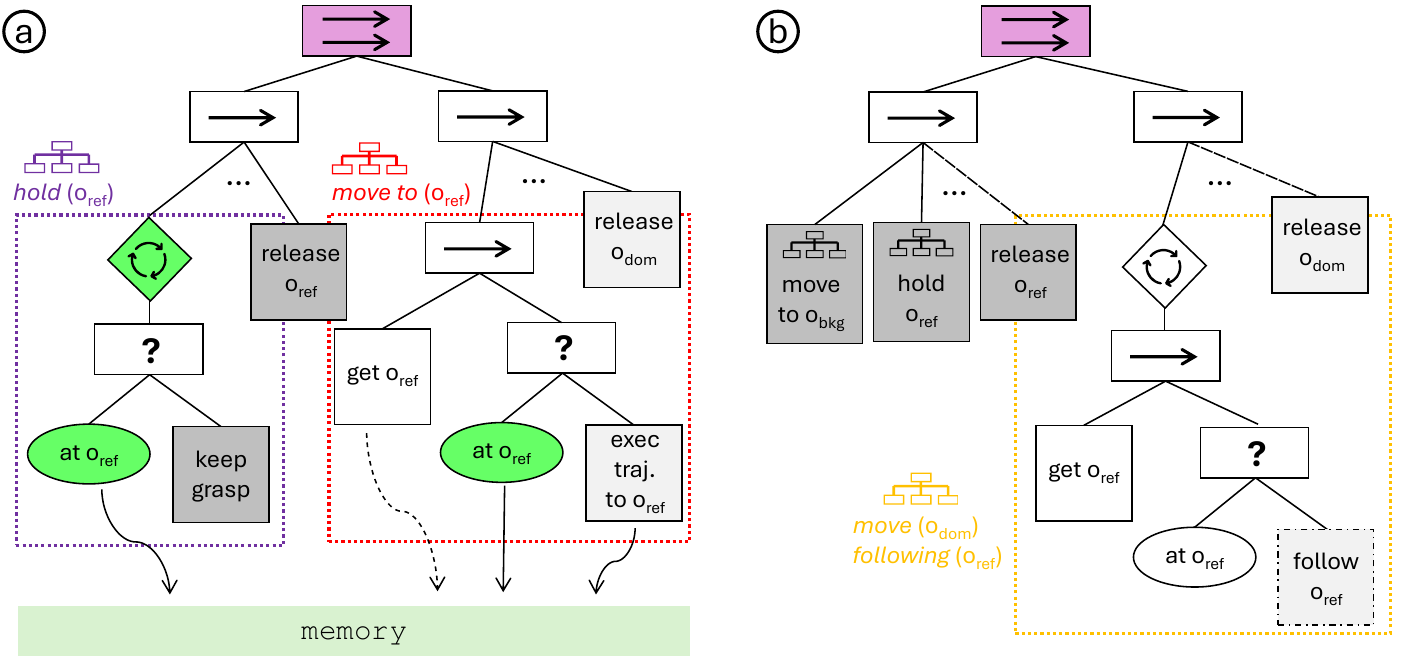}
\end{adjustwidth}
\vspace{-0.4cm}
\caption{Subtree for a \textit{sequential} dual-arm activity, when the reference object is held stationary (a) and when is moved (b).}
\vspace{-0.4cm}
\label{fig:sequential}
\end{figure}

In \textit{sequential} bimanual activities, such as holding a pot with one hand and putting the lid on it with the other, $sBT_{seq}$ has the structure shown in Fig. \ref{fig:sequential} (a). The parent is a parallel node, with two $sBT_{one-arm}$ as children, but unlike $sBT_{uncoo}$, these subtrees are interdependent. Indeed, the $sBT_{one-arm}$ that instructs the arm to handle the dominant object $o_{dom}$ (e.g., the lid) includes $sBT_{move \text{ to } o_{ref}}$, which computes and executes the trajectory towards the reference object $o_{ref}$ (e.g., the pot) which is held by the other arm. 
Thus, the other $sBT_{one-arm}$ has $sBT_{hold \text{ } o_{ref}}$ to hold $o_{ref}$. A decorator node of type \textit{KeepRunningUntilSuccess} ensures that the \textit{keep grasp} command is repeatedly executed until the condition \textit{at $o_{ref}$} is met, which occurs once the trajectory of $o_{dom}$ is complete. 
Moreover, the main difference with $sBT_{uncoo}$ is that, although the two arms operate simultaneously thanks to the parallel node, they have the same condition for completing their activities, allowing temporal coordination of the two arms. When such a condition is met, the \textit{sequential} activity could terminate with a simultaneous release command, making both end-effectors ready for new instructions.
If $o_{ref}$ is not just held stationary but moved towards a target object $o_{bkg}$, the activity of the arm handling $o_{dom}$ would be represented by $sBT_{move \text{ } o_{dom} \text{ } following \text{ } o_{ref}}$, as illustrated in Fig. \ref{fig:sequential} (b). The decorator node ensures that the detection of $o_{ref}$ and the adjustment of $o_{dom}$ trajectory towards $o_{ref}$ occur in a continuous loop until \textit{at $o_{ref}$} is satisfied.

\vspace{-0.2cm}
\section{Experiments}
In our experimental campaign, we evaluated the performance of our pipeline in recognizing arms coordination mode, extracting primitive sequences, and generating effective BTs for successful task replication. 
We first processed multi-subject video demonstrations from our open-source dataset. These tasks, designed to highlight the defined coordination modes, were conducted in a controlled environment with subjects following instructions. 
Next, we tested the method on an external publicly available dataset to demonstrate its effectiveness on non-customized tasks. 
Additionally, we compared our bimanual task representation with that of \cite{dreher2020learning}.
\vspace{-0.4cm}

\subsection{Pipeline Assessment using HANDSOME Dataset}
We tested our framework by processing videos of bimanual activities from our dataset HANDSOME\footnote{The multimedia attachment related to this experimental phase is in the additional materials and online at \href{https://youtu.be/RZngyFtoOoE}{https://youtu.be/RZngyFtoOoE}.}. We recorded demonstrations from ten subjects using an Intel RealSense D435i camera positioned above the workspace. A marker-based system was employed to accurately detect the $6$D poses of both objects and hands over time\footnote{All details regarding the data recording procedure, the adopted objects, and the subjects involved are included in the open-source documentation at \href{https://doi.org/10.5281/zenodo.13846970}{doi.org/10.5281/zenodo.13846970}.}. 
This dataset comprehends bimanual tasks performed in two contexts (kitchen and workshop), using objects typical of each scenario. Table \ref{tab:tasks} shows the instructions given to the participants for carrying out the tasks.
We processed all 150\footnote{Each participant completed $3$ different tasks, each repeated $5$ times.} collected videos, analyzing the quality of both the segmentation and the generated BT. Since our method is one-shot, meaning we can generate an execution plan from a single demonstration, we selected one video per task and executed the generated BT using our dual-arm robotic system, consisting of two Franka Emika Panda robots placed side by side (named \textit{franka A} and \textit{franka B}). 
For the replica, we used an Intel RealSense D435i camera above the workspace to retrieve the initial object poses. 
The software architecture, developed in Python, ran on Ubuntu 20.04 on an Alienware laptop with an Intel i7 processor, NVIDIA RTX 2080 GPU, and 32 GB RAM.

\begin{table}[t]
\caption{Bimanual Activities in HANDSOME Dataset}
\label{tab:tasks}
\centering
\begin{tabular}{|c||c|c|}
\hline
 \hspace{-0.2cm} \textbf{Task} \hspace{-0.2cm} & \textbf{Kitchen} & \textbf{Workshop}\\
\hline
\scriptsize\textbf{1} & \makecell{\scriptsize put \textit{cup$_1$} on \textit{plate$_1$} then \\ \scriptsize put \textit{cup$_2$} on \textit{plate$_2$}} & \makecell{\scriptsize simultaneously put \textit{profile$_{B1}$} on \textit{blue ink} \\ \scriptsize and \textit{profile$_C$} on \textit{yellow ink}} \\
\hline
\scriptsize\textbf{2} & \makecell{\scriptsize co-transport \textit{pan} on \textit{cooker} 
} & \makecell{\scriptsize co-transport \textit{box} on \textit{workstation} 
} \\
\hline
\scriptsize\textbf{3} & \makecell{\scriptsize put \textit{pan} on \textit{cooker} with one \\ \scriptsize hand and hold it; put \textit{cover}\\ \scriptsize on \textit{pan} with the other hand} & 
\makecell{\scriptsize put \textit{joint} on \textit{workstation} with one hand \\ \scriptsize  and hold it; grasp \textit{profile$_A$} with \\ \scriptsize the other and assemble it with \textit{joint}; \\ \scriptsize co-transport the assembly on \textit{scale}} \\
\hline
\end{tabular}
\vspace{-0.4cm}
\end{table}

\vspace{-0.2cm}
\subsection{Pipeline Assessment and Comparison with a State-of-the-art approach using The KIT Bimanual Dataset}
\vspace{-0.1cm}
To measure the effectiveness of our algorithm in processing even non-customized tasks, we additionally validated it by processing data from an external dataset of bimanual activities (\href{https://bimanual-actions.humanoids.kit.edu/original_dataset}{The KIT Bimanual Dataset}); we tested the quality of the retrieved segmentation and the generated BTs. 
We selected the pouring task, which was performed $10$ times by $6$ subjects, taking only those repetitions where the scene contained just one instance for each object category to ensure consistent object tracking. 
In addition, we compared the pouring task representation resulting from our segmentation with the one obtained in \cite{dreher2020learning} using the same dataset. 

\vspace{-0.2cm}
\section{Results}
\vspace{-0.1cm}
\subsection{Pipeline Assessment using HANDSOME Dataset}

\begin{figure} [t]
\centering
\includegraphics[width=\linewidth]{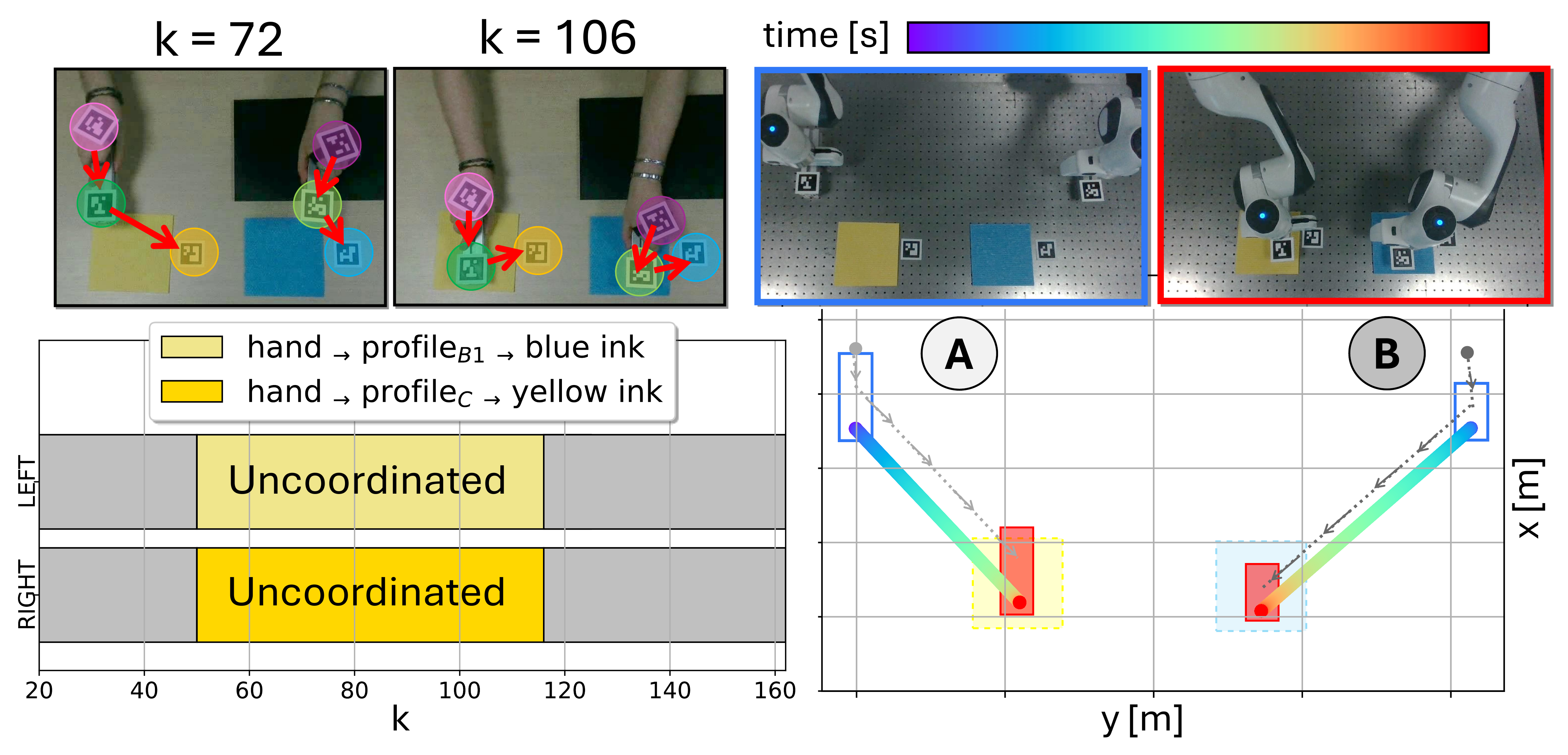}
\vspace{-0.8cm}
\caption{(Left) Keyframes and segmentation of Task $1$ in the workshop context, and (right) the robot replica. Blue and yellow squares represent the two ink sources. Profiles are drawn in blue at the start of task execution and in red at the end, with the same colors used as frame contours to illustrate object and robotic arm configurations at those moments. Gray dots indicate the end effectors' homing positions (light gray for \textit{franka A} and dark gray for \textit{franka B}), with dashed gray lines for their trajectories and colored trails for profiles.}
\label{fig:b1w}
\vspace{-0.3cm}
\end{figure}

For all demonstrations of Task $1$ in the workshop context, the \textit{uncoordinated} mode was recognized as expected, due to the two hands operating at the same time. 
Fig. \ref{fig:b1w} shows the video processing results (left) and robotic replica (right) of one performance. 
The participant simulated marking two aluminum profiles ($B1$ and $C$) simultaneously with different inks (blue and yellow). 
The scene graphs highlighted interactions between each hand and the profiles, and between the profiles and the corresponding ink ($k=72$, $k=106$), with no shared nodes between graphs $G_R$ and $G_L$ (topology $\alpha$) defining an \textit{uncoordinated} activity. 
The generated BT instructed the two robots to parallel perform pick-and-place activities, as shown by the two rainbow lines depicting profile trajectories over time (see Fig. \ref{fig:b1w}(right)). 

Contrarily, Task $1$ in the kitchen scenario was performed by first moving one cup with one hand and then the other cup with the other hand. Therefore, the two activities were recognized as \textit{one arm}, and the BT featured a sequence of two similar $sBT_{one-arm}$, commanding the relocation of one cup and, once completed, the relocation of the other.

\begin{figure} [t]
\centering
\includegraphics[width=\linewidth]{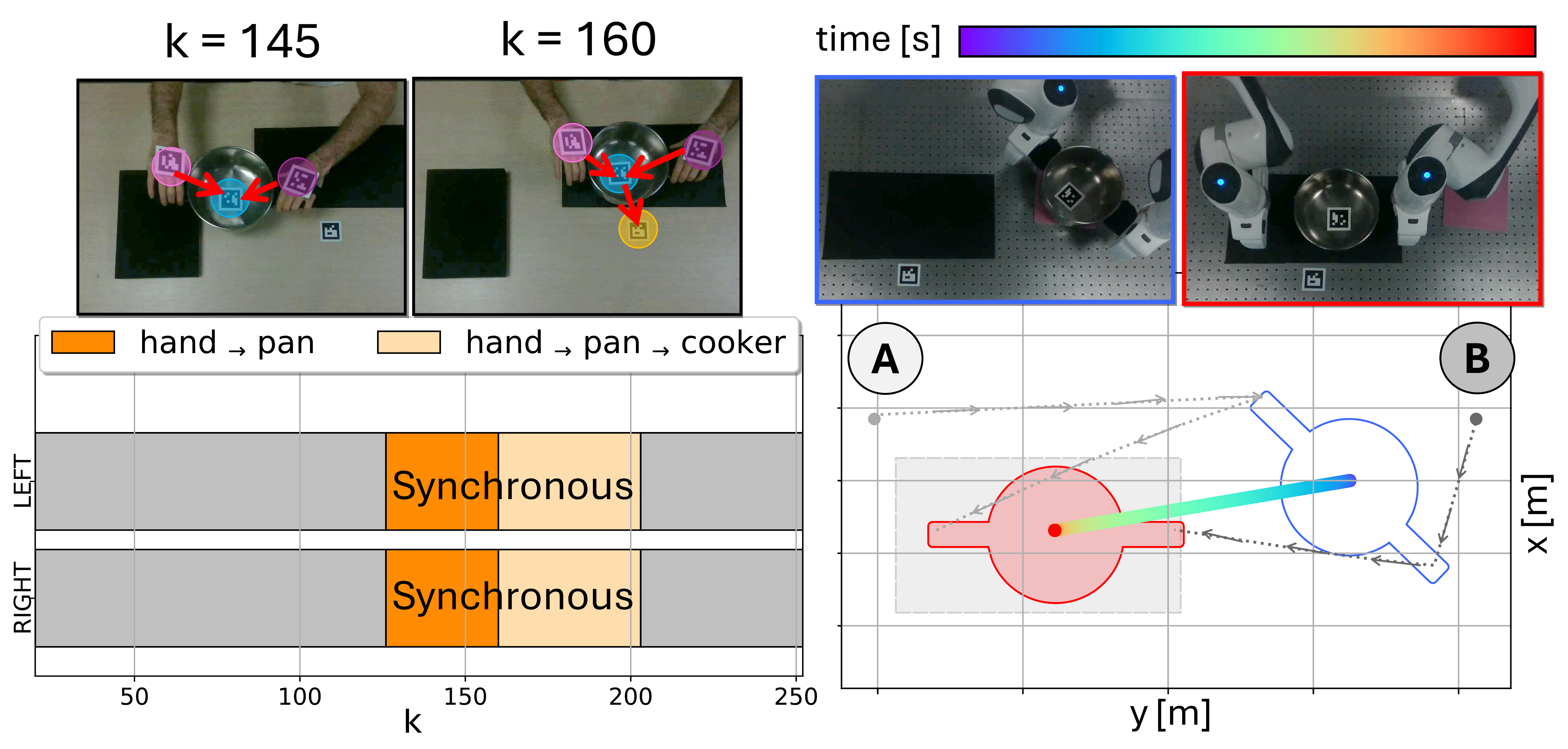}
\vspace{-0.8cm}
\caption{(left) Keyframes and segmentation of Task $2$ in the kitchen context and (right) robot replica. The gray rectangle is the cooker, while the pan is in blue at the start of the execution and in red at the end.}
\label{fig:b2k}
\vspace{-0.6cm}
\end{figure}

Across both contexts, the algorithm identified the co-transport activity in Task $2$ as \textit{synchronous} in all the demonstrations. 
Processing results of one performance in the kitchen context are illustrated in Fig. \ref{fig:b2k}(left). 
The subject used both hands to grasp the pan by the handles and place it on the induction cooker. 
The graphs $G_R$ and $G_L$ were merged into a single $G_B$ since both hands interacted with the same object, the pan. Initially, $G_B$ only included $RO$ and $LO$ ($k=145$), and later the pan-cooker interaction was detected ($k=160$). Two distinct IUs, representing pan transport and positioning, were identified (highlighted with different orange shades). Since all $G_B$ shared the same topology ($\gamma$), the IUs were labeled as \textit{synchronous}. 
The robot replica results in Fig. \ref{fig:b2k}(right) show both arms independently grasping the handles and moving the pan to the cooker via a coordinated trajectory. 
The BT guiding this replica consisted of a single $sBT_{sync}$ that instructed the two arms to grasp the pan, move together, and then release it. In some instances, $RO$ was detected before $LO$ (and vice versa), resulting in an initial $sBT_{one-arm}$ commanding one arm to reach and grasp the pan, while the other instructions remained in $sBT_{sync}$. Similarly, in a few demonstrations, $RO$ and $LO$ did not end simultaneously, resulting in a final $sBT_{one-arm}$ including the release command for one arm. Despite slight variations in BTs, the system consistently completed the task successfully.

\begin{figure} [t]
\centering
\vspace{-0.1cm}
\includegraphics[width=\linewidth]{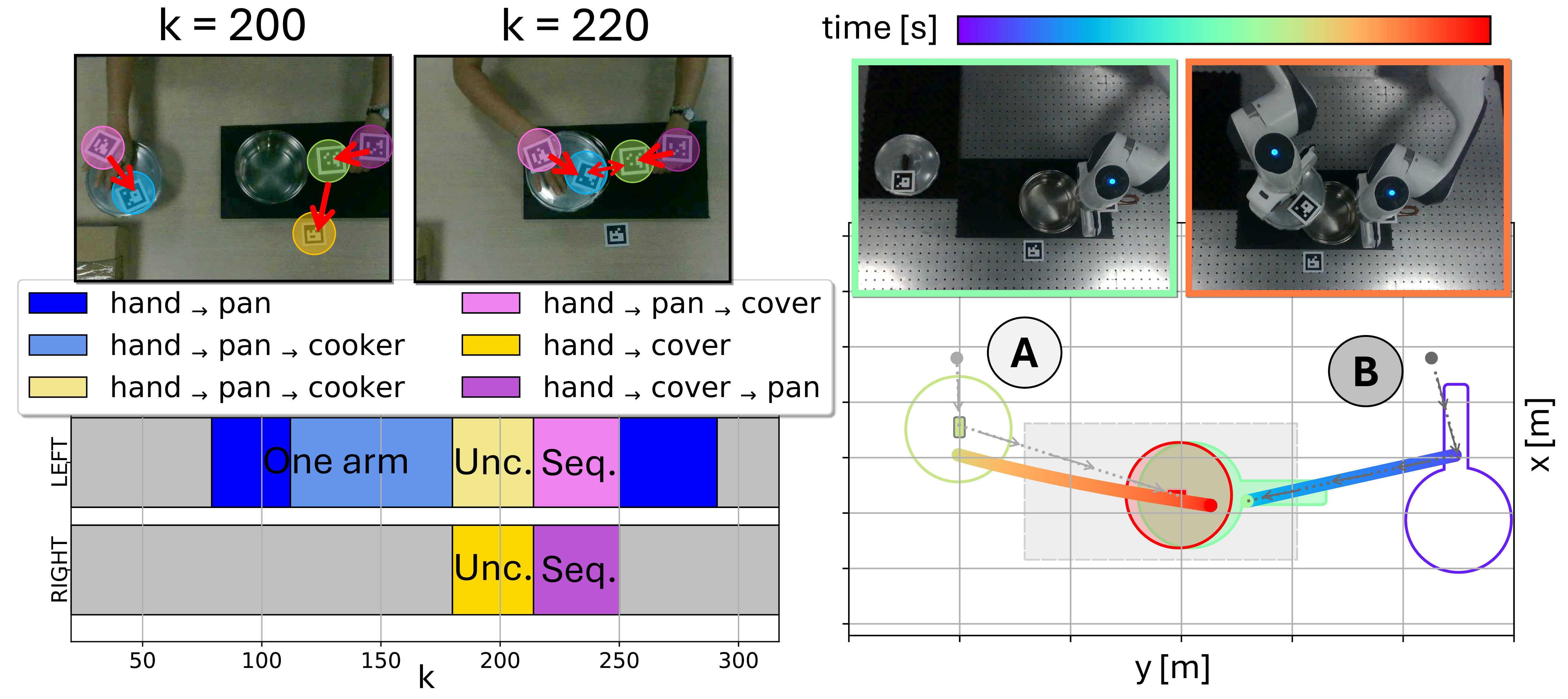}
\vspace{-0.7cm}
\caption{(Left) Keyframes and segmentation of Task $3$ in the kitchen context, and (right) the robot replica. Frame contour colors indicate object configurations: aquamarine for when the pan is on the cooker, and orange for when the cover reaches the pan. The gray rectangle represents the cooker. The pan appears in violet at the start of the task and in aquamarine once on the cooker. The cover is shown in green at the start of manipulation and in red at the end.}
\label{fig:b3k}
\vspace{-0.2cm}
\end{figure}

For all demonstrations of Task $3$ in the kitchen context, the segmentation is the one shown in Fig. \ref{fig:b3k}(left), as the participants were instructed on how to perform the task. First, the left hand grasped and moved the pan on the cooker, generating two IUs labeled as \textit{one arm} since the right hand was stationary. When the right hand started manipulating the cover and the left hand maintained the grip on the pan, both $G_R$ and $G_L$ were created (see $k=200$), $G_B$ assumed topology $\alpha$, and this IU was labeled \textit{uncoordinated}. As the cover approached the pan, it resulted in a $G_B$ of topology $\eta$ ($k=220$), and the IU was recognized as \textit{sequential}, with the pan as $o_{ref}$ and the cover as $o_{dom}$, since $MI_{R, cover}>MI_{L, pan}$. 
The final IU was labeled \textit{one arm} after the right hand released the cover and moved away, leaving only $G_L$.
During replication (see Fig. \ref{fig:b3k}(right)), \textit{franka B} acted as the left hand by moving the pan on the cooker and keeping it grasped while \textit{franka A} placed the cover on top.

In the workshop variant of this task, the \textit{sequential} activity was detected properly during the assembly of the two pieces. 
When the two hands moved the assembly together onto the scale, the coordination mode correctly switched to \textit{synchronous}. 

\vspace{-0.3cm}
\subsection{Results on The KIT Bimanual Dataset}
\vspace{-0.1cm}
\subsubsection{Pipeline Assessment}


\begin{figure} [t]
\centering
\includegraphics[width=0.8\linewidth]{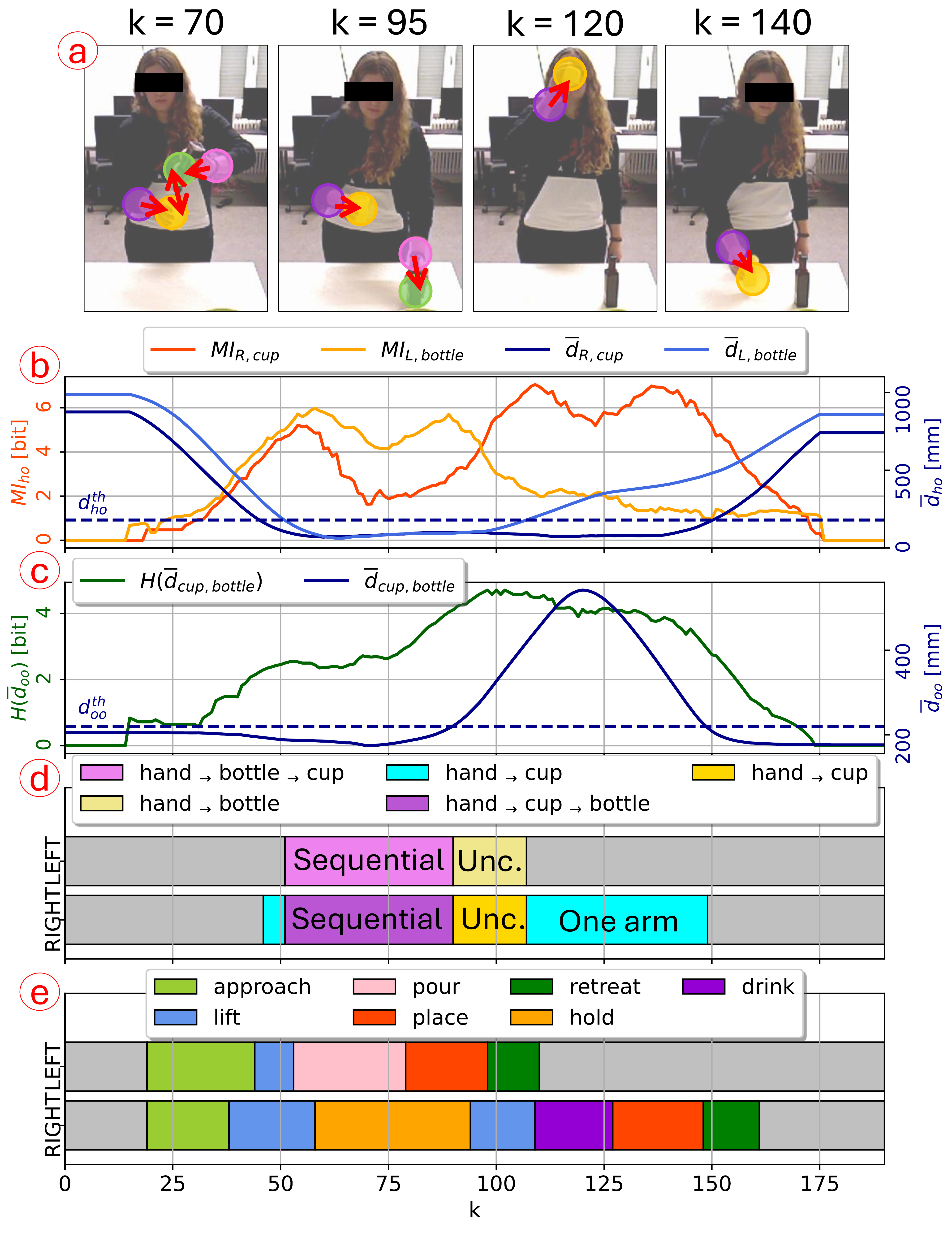}
\vspace{-0.3cm}
\caption{(a) Keyframes of a pouring activity from The KIT Bimanual dataset; (b) trend of mutual information and average distance between each hand and the manipulated object; (c) trend of cup-bottle average distance and entropy; (d) segmentation obtained by processing the demonstration with our method and (e) with the method proposed in \cite{dreher2020learning}.}
\vspace{-0.5cm}
\label{fig:pour}
\end{figure}

\begin{figure} [t]
\centering
\includegraphics[width=0.82\linewidth]{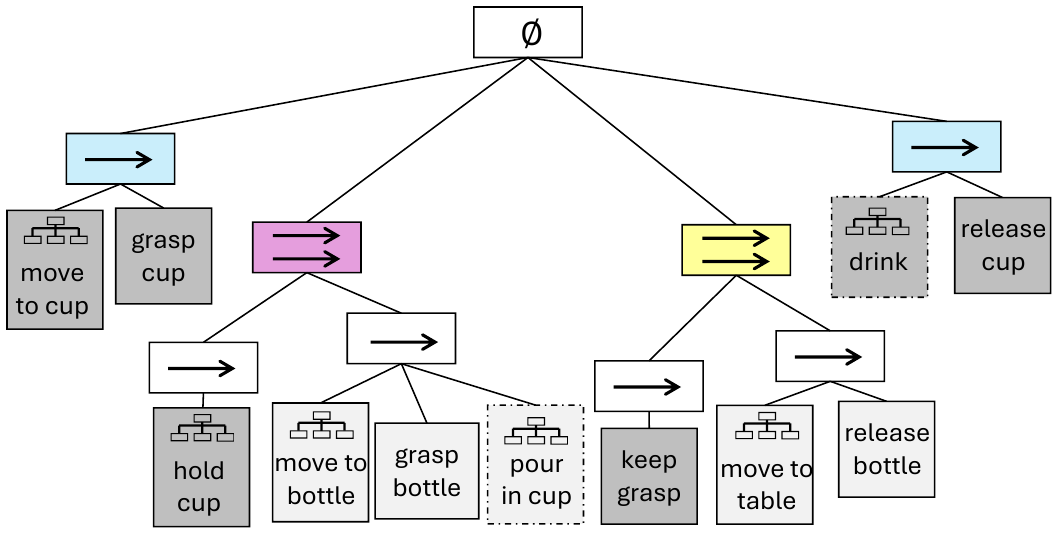}
\vspace{-0.3cm}
\caption{The BT generated based on our segmentation of pouring activity. The colored control nodes initiate a subtree with a new coordination mode.}
\label{fig:bt_pour}
\vspace{-0.5cm}
\end{figure}

We selected one demonstration from subject $2$ as a representative execution (see Fig. \ref{fig:pour}(a)) and presented the segmentation results in Fig. \ref{fig:pour}(d).  
During the pouring phase, $G_B$ had the topology shown for $k=70$, and the activity was classified as \textit{sequential}, with the bottle as $o_{dom}$ and cup as $o_{ref}$, since $MI_{L,bottle} > MI_{R, cup}$. While the left hand placed the bottle on the table, the right hand held the cup: $G_R$ and $G_L$ remained separate, as seen at $k=95$, labeling the IU as \textit{uncoordinated}. In the drinking phase, only the right hand was active (see $k=120$), then the activity required \textit{one arm} only.
Fig. \ref{fig:bt_pour} illustrates the BT generated from our segmentation. 
The \textit{sequential} IU was mapped into a couple of interdependent subtrees triggered by the parallel node in violet, while the \textit{uncoordinated} IU into two independent subtrees called by the parallel node in yellow. The two \textit{one arm} IUs became two $sBT_{one-arm}$.
%
Since each subject performed the pouring activity in their own way, we obtained diverse segmentation results and BTs. Specifically, we observed that: (i) the hand manipulating the bottle changed, but we were always able to recognize the bottle as $o_{dom}$; (ii) sometimes the bottle was not released once relocated, resulting in a longer \textit{uncoordinated} activity including the drinking motion; (iii) in some videos, subjects started lifting both cup and bottle at the same time, which resulted in an \textit{uncoordinated} IU before the \textit{sequential} one; (iv) in some repetitions the person switched the cup from one hand to the other to perform both pouring and drinking with the same hand. Our system does not handle this handover between hands, but we could address it by adding instructions for releasing the cup with one arm and grasping it with the other.
Note that we encountered some issues in running our pipeline when positional data were very noisy due to the bad detection of the hands or when more than one instance of the same object was erroneously detected.

\subsubsection{Comparison with Bimanual Task Representation in \cite{dreher2020learning}}
Clear differences can be observed between our segmentation (Fig. \ref{fig:pour}(d)) and the one obtained using the method from \cite{dreher2020learning} (Fig. \ref{fig:pour}(e)). 
The latter is more descriptive than ours from a semantic point of view and effectively communicates the activity content to a person, using natural language labels (like \textit{approach}, \textit{pour}, \textit{drink}). 
However, these details do not aid in the automatic generation of a plan; rather, they reduce its interpretability for a robot. Additionally, \cite{dreher2020learning} returns actions for each hand separately while ignoring whether and how the two hands are coordinating.
Based on this representation, the generation of a plan involving specific coordination policies between the two arms becomes more complex, as in the pouring action. In contrast, our method identified the hand coordination strategy employed during the demonstration and automatically created a plan to replicate it, shown in Fig. \ref{fig:bt_pour}. 
Fig. \ref{fig:pour}(b) and (c) also show the information-theoretic metrics calculated throughout the human performance to drive interaction detection and task segmentation, as described in \cite{merlo2024exploiting}\footnote{Thresholds $d_{ho}^{th}$ and $d_{oo}^{th}$ define the minimum distance for interaction, based on prior studies \cite{diehl2021automated}.}. Specifically, $MI_{R, cup}$, $MI_{L, bottle}$, and the average distances $\overline{d}_{R, cup}$ and $\overline{d}_{L, bottle}$ were used to guide $HO$ detection. Meanwhile, $\overline{d}_{cup, bottle}$ and its entropy $H(\overline{d}_{cup, bottle})$ guided $OO$ detection.
By analyzing $MI_{R, cup}$ and $MI_{L, bottle}$ trends, we can obtain low-level information about the motion patterns of the manipulated objects. For instance, we can distinguish between broad movements, like relocating an object, and precise, confined ones, like pouring.

\vspace{-0.2cm}
\section{Discussion and Conclusion}
\vspace{-0.2cm}
In this paper, we presented an intuitive programming method for dual-arm robotic systems based on single demonstrations of bimanual activities recorded using standard RGB cameras. By leveraging Shannon's information theory and the properties of scene graphs, the video processing was able to generate a high-level representation of the demonstrated task, identify the hand coordination strategies, and 
automatically generate a centralized execution plan. 
The experimental campaign was conducted by processing video from two sources: an internal open-source dataset, including multi-subjects demonstrations with real-world objects and realistic scenarios, and an external dataset. Our algorithm also performed well when applied to recordings from the external dataset, where the field of view and the system used to detect objects and hands varied from ours. This indicates that the proposed method is flexible and can accommodate different perception modules, provided that the position and orientation of each scene element are available. Additionally, the results proved the robustness of the proposed method in handling positional data from a markerless perception system, which tends to be less accurate.
Currently, the system cannot replicate complex movements such as pouring, but the plan is prepared to instruct the system to perform these movements upon integration of trajectory learning algorithms. However, it should be recalled that the focus of our algorithm is to capture a high-level representation of the task and retrieve an appropriate robotic execution plan, while the specific motions can be integrated later based on the platform and object features.

While generating a robot plan from a single human demonstration is a strength, a natural evolution of the framework would involve analyzing multiple demonstrations to optimize the plan, such as reducing instructions or execution time. Additionally, by evaluating the platform’s capabilities, we could identify bimanual tasks that could be performed with a single robot, optimizing resources.
Finally, to increase versatility, future developments of our framework will allow users to engage directly and intuitively with the system to adapt and customize the generated plans through multi-modal inputs.


\vspace{-0.2cm}


\bibliographystyle{IEEEtran}
\bibliography{biblio}

\end{document}